%% file: auto-corex.tex
\documentclass{article}
\usepackage{microtype}
\usepackage{graphicx}
\usepackage{booktabs} 
\usepackage{hyperref}

\usepackage{amsfonts}
\usepackage{amsmath}
\usepackage{amsthm}
\usepackage{graphicx}
\usepackage{subcaption}
\newcommand{\vx}{{\mathbf x}}

\newcommand{\vz}{{\mathbf z}}

\newcommand{\be}{\begin{eqnarray} \begin{aligned}}
\newcommand{\ee}{\end{aligned} \end{eqnarray}}
\newcommand{\benn}{\begin{eqnarray*} \begin{aligned}}
\newcommand{\eenn}{\end{aligned} \end{eqnarray*} }

\newtheorem*{theorem*}{Theorem}

\usepackage[accepted]{icml2018}
\newcommand{\aram}[1]{ \textcolor{red}{[#1]}}

\usepackage{enumitem}

\icmltitlerunning{Auto-Encoding Total Correlation Explanation}

\begin{document}
\twocolumn[
\icmltitle{Auto-Encoding Total Correlation Explanation}
\icmlsetsymbol{equal}{*}
\begin{icmlauthorlist}
\icmlauthor{Shuyang Gao}{isi}
\icmlauthor{Rob Brekelmans}{isi}
\icmlauthor{Greg Ver Steeg}{isi}
\icmlauthor{Aram Galstyan}{isi}
\end{icmlauthorlist}
\icmlaffiliation{isi}{Information Sciences Institute, University of Southern California, Marina Del Rey, California, USA}
\icmlcorrespondingauthor{Shuyang Gao}{sgao@isi.edu}
\icmlkeywords{Machine Learning, ICML}
\vskip 0.3in
 ]
\printAffiliationsAndNotice{} 
\begin{abstract}
Advances in unsupervised learning enable reconstruction and generation of samples from complex distributions, but this success is marred by the inscrutability of the representations learned. 
We propose an information-theoretic approach to characterizing disentanglement and dependence in representation learning using multivariate mutual information, also called total correlation. The principle of total Cor-relation Ex-planation (CorEx) has motivated successful unsupervised learning applications across a variety of domains, but under some restrictive assumptions. Here we relax those restrictions by introducing a flexible variational lower bound to CorEx. Surprisingly, we find that this lower bound is equivalent to the one in variational autoencoders (VAE) under certain conditions. This information-theoretic view of VAE deepens our understanding of hierarchical VAE and motivates a new algorithm, AnchorVAE, that makes latent codes more interpretable through information maximization and enables generation of richer and more realistic samples.
\end{abstract}

\input{sections/intro}
\input{sections/background}

\input{sections/method}

\input{sections/connection}
\input{sections/results}

\input{sections/conclusion-related}

\newpage
\bibliography{sections/shuyang_bib,sections/gversteeg_bibdesk_no_shuyang_duplicates} 
\bibliographystyle{icml2018}

\end{document}

%% file: sections/intro.tex
\section{Introduction}
\label{sec:intro}
Learning representations from data without labels has become increasingly important to solving some of the most crucial problems in machine learning---including tasks in image, language, speech, etc.~\cite{bengio2013representation}. Complex models, such as deep neural networks, have been successfully applied to generative modeling with high-dimensional data. From these methods we can either infer hidden representations with variational autoencoders (VAE)~\cite{kingma2013auto, rezende2014stochastic} or generate new samples with VAE or generative adversarial networks (GAN) ~\cite{goodfellow2014generative}.

Building on these successes, an explosive amount of recent effort has focused on \textit{interpreting} learned representations, which could have significant implications for subsequent tasks. Methods like InfoGAN~\cite{chen2016infogan} and $\beta$-VAE~\cite{higgins2016beta} are able to learn \textit{disentangled} and \textit{interpretable} representations in a completely unsupervised fashion. 
Information theory provides a natural framework for understanding representation learning and continues to generate new insights~\cite{alemi2016deep, shwartz2017opening, achille2018information, saxe2018info}.

In this paper we discuss the problem of learning \textit{disentangled} and \textit{interpretable} representations in a purely information-theoretic way. Instead of making assumptions about the data generating process at the beginning, we consider the question of how informative the underlying latent variable $\vz$ is about the original data variable $\vx$. We would like $\vz$ to be as informative as possible about the \textit{relationships} in $\vx$ while remaining as disentangled as possible in the sense of statistical independence. This principle has been previously proposed as Cor-relation Ex-planation (CorEx)~\cite{ver2014discovering, steeg2017unsupervised}.  By optimizing appropriate information-theoretic measures, CorEx defines not only an \textit{informative} representation but also a \textit{disentangled} one, thus eliciting a natural comparison to the recent literature on interpretable machine learning. However, computing the CorEx objective can be challenging, and previous studies have been restricted to cases where random variables are either discrete~\cite{ver2014discovering}, or Gaussian~\cite{steeg2017low}. 

Our key contributions are as follows: 
\vspace{-1mm}
\begin{itemize}[noitemsep,topsep=0pt]
\item We construct a variational lower bound to the CorEx objective and optimize the bound with deep neural networks. Surprisingly, we find that under standard assumptions, the lower bound for CorEx shares the same mathematical form as the evidence lower bound (ELBO) used in VAE, suggesting that CorEx provides a dual information-theoretic perspective on representations learned by VAE. 
\item Going beyond the standard scenario to hierarchical VAEs or deep Gaussian latent models (DLGM)~\cite{rezende2014stochastic}, we demonstrate that CorEx provides new insight into measuring how representations become progressively more disentangled at subsequent layers. In addition, the CorEx objective can be naturally decomposed into two sets of mutual information terms with an interpretation as an unsupervised information bottleneck.
\item Inspired by this formulation, we propose to make some latent factors more interpretable by reweighting terms in the objective to make certain parts of the latent code uniquely informative about the inputs (instead of adding new terms to the objective, as in InfoGAN~\cite{chen2016infogan}).
\item Finally, we show that by sampling each latent code $\vz_i$ from the encoding distribution $p(\vz_i) = \int_{\vx} { p(\vz_i|\vx)p(\vx) d\vx}$ instead of the standard Gaussian prior in VAE, we can generate richer and more realistic samples than VAE even under the same network model. 
\end{itemize}

We first review some basic information-theoretic quantities in Sec.~\ref{sec:back}, then introduce the total correlation explanation (CorEx) learning framework in Sec.~\ref{sec:corex}. In Sec.~\ref{sec:method} we derive the variational lower bound of the CorEx objective and demonstrate a connection with VAE in Sec.~\ref{sec:connection}. This connection sheds light on some new applications of VAE, which we will describe in Sec.~\ref{sec:app}. We discuss related work in Sec.~\ref{sec:related} and conclude our paper in Sec.~\ref{sec:con}. 

%% file: sections/background.tex
\section{Information Theory Background}\label{sec:back}
Let $\vx = (\vx_1, \vx_2, ..., \vx_d)$ denote a $d$-dimensional random variable whose probability density function is $p(\vx)$. Shannon differential entropy~\cite{cover} is defined in the usual way as $H(\vx) = -\mathbb{E}_\vx\left[\log p(\vx)\right]$. Let $\vz = (\vz_1, \vz_2, ..., \vz_m)$ denote an $m$-dimensional random variable whose probability density function is $p(\vz)$. Then mutual information between two random variables, $\vx$ and $\vz$, is defined as $I(\vx:\vz)=H(\vx)+H(\vz)-H(\vx,\vz)$. Mutual information can also be viewed as the reduction in uncertainty about one variable given another variable---i.e.,  $I(\vx:\vz)=H(\vx)-H(\vx|\vz)=H(\vz)-H(\vz|\vx)$.

A measure of multivariate mutual information called \textit{total correlation}~\cite{watanabe} or \textit{multi-information}~\cite{multiinformation} is defined as follows:
\begin{equation}
\label{eq:tc}
TC\left(\vx\right) = \sum_{i=1}^{d}H\left(\vx_i\right) - H\left(\vx\right) 
= D_{KL}\left(p(\vx)||\prod_{i=1}^{d}p(\vx_i)\right) 
\end{equation}
Note that $D_{KL}\left(\cdot\right)$ denotes the \textit{Kullback-Leibler} divergence in Eq.~\ref{eq:tc}. Intuitively, $TC(\vx)$ captures the total dependence across all the dimensions of $\vx$ and is zero if and only if all $\vx_i$ are independent. Total correlation or statistical independence is often used to characterize \textit{disentanglement} in recent literature on learning representations \cite{nice,achille2017emergence}.

The conditional total correlation of $\vx$, after observing some latent variable $\vz$, is defined as follows,
\begin{equation}
\label{eq:condtc}
\begin{split}
TC\left(\vx|\vz\right) =& \sum_{i=1}^{d}H\left(\vx_i|\vz\right) - H\left(\vx|\vz\right)  \\
=& D_{KL}\left(p(\vx|\vz)||\prod_{i=1}^{d}p(\vx_i|\vz)\right) 
\end{split}
\end{equation}
We define a measure of \textit{informativeness} of latent variable $\vz$ about the dependence among the observed variables $\vx$ by quantifying how total correlation is reduced after conditioning on some latent factor $\vz$; i.e., 
\begin{equation} \label{eq:corex}
TC\left(\vx;\vz\right) = TC(\vx)-TC(\vx|\vz)
\end{equation}
In Eq.~\ref{eq:corex}, we can see that $TC\left(\vx;\vz\right)$ is maximized if and only if the conditional distribution $p(\vx|\vz)$ factorizes, in which case we can interpret $\vz$ as capturing the information about \textit{common causes} across all $\vx_i$. 

\section{Total Correlation Explanation Representation Learning} \label{sec:corex}
In a typical unsupervised setting like VAE, we assume a generative model where $\vx$ is a function of a latent variable $\vz$, and we then maximize the log likelihood of $\vx$ under this model. From a CorEx perspective, the situation is reversed. We let $\vz$ be some stochastic function of $\vx$ parameterized by $\theta$, i.e., $p_{\theta}(\vz|\vx)$. Then we seek a  joint distribution $p_{\theta}(\vx, \vz) = p_{\theta}(\vz|\vx)p(\vx)$, where $p(\vx)$ is the underlying true data distribution that maximizes the following objective:
\begin{equation} \label{eq:corex_lb}
\begin{split}
\mathcal{L}(\theta;\vx) &=  \underbrace{TC_{\theta}(\vx;\vz)}_{\text{informativeness}} - \underbrace{TC_{\theta}(\vz)}_{\text{(dis)entanglement}} \\
&=TC(\vx) - TC_\theta(\vx|\vz) - TC_\theta(\vz)
\end{split}
\end{equation}
In Eq.~\ref{eq:corex_lb}, $TC_{\theta}(\vx;\vz)$ corresponds to the amount of correlation that is explained by $\vz$ as defined in Eq.~\ref{eq:corex}, and $TC_{\theta}(\vz)$ quantifies the dependence among the latent variables $\vz$. 

\begin{figure}[ht]
	\centering
	\includegraphics[width=0.8\columnwidth]{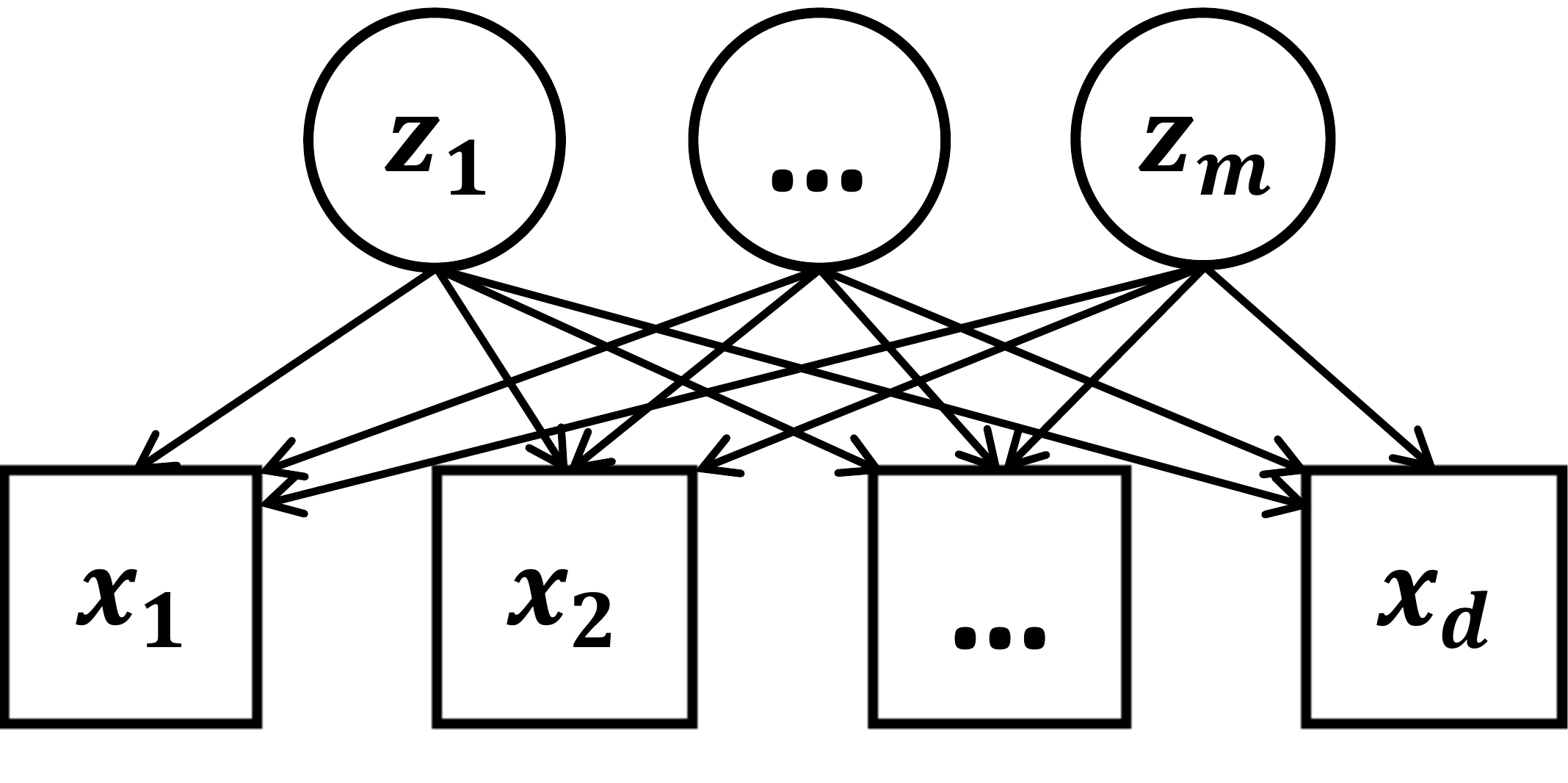}
	\caption{The graphical model for $p_{\theta^*}(\vx,\vz)$ assuming $p_{\theta^*}(\vz|\vx)$ achieves the global maximum in Eq.~\ref{eq:corex_lb}. In this model, all $\vx_i$ are factorized conditioned on $\vz$, and all $\vz_i$ are independent. }\vskip -0.1in
	\label{fig:corex_model}
\end{figure}

By non-negativity of total correlation, Eq.~\ref{eq:corex_lb} naturally forms a lower bound on $TC(\vx)$; i.e., $TC(\vx) \ge \mathcal{L}(\theta;\vx)$ for any $\theta$. Therefore, the global maximum of Eq.~\ref{eq:corex_lb} occurs at $TC(\vx)$, in which case $TC_{\theta^*}(\vx|\vz) = TC_{\theta^*}(\vz)\equiv 0$ and  $\vz$ can be exactly interpreted as a generative model where $\vz$ are independent random variables that generate $\vx$, as shown in Fig.~\ref{fig:corex_model}.

Notice that the term $TC_\theta(\vx;\vz)$ is a bit different from the classical definition of informativeness using mutual information $I_\theta(\vx;\vz)$~\cite{linsker}.  In fact, after combining the entropy terms in Eq.~\ref{eq:tc} and ~\ref{eq:condtc}, the following equation holds~\cite{ver2015maximally}:
\begin{equation} \label{eq:corex_expand}
TC_{\theta}(\vx;\vz) = \sum_{i=1}^d I_\theta(\vx_i;\vz) - I_\theta(\vx;\vz)
\end{equation}
The term $TC_\theta(\vx;\vz)$ in Eq.~\ref{eq:corex_lb} can be seen as finding a minimal latent representation $\vz$ which, after conditioning, disentangles $\vx$.  When stacking hidden variable layers in Sec. ~\ref{sec:app}, we will see that this condition can lead to interpretable features by forcing intermediate layers to be explained by higher layers under a factorized model.

\paragraph{Informativeness vs.   Disentanglement} If we only consider the informativeness term $TC_{\theta}(\vx;\vz)$ as in the objective, a naive solution to this problem would be just setting $\vz=\vx$. To avoid this, we also want the latent variables $\vz$ to be as disentangled as possible, corresponding to the $TC(\vz)$ term encouraging independence. In other words, the objective in Eq.~\ref{eq:corex_lb} is trying to find $\vz$, so that $\vz$ not only disentangles $\vx$ as much as possible, but is itself as disentangled as possible.

%% file: sections/method.tex
\section{Optimization} \label{sec:method}
We first focus on optimizing the objective function defined by Eq.~\ref{eq:corex_lb}. The extension to the multi-layer (hierarchical) case is presented in the next section. 

By using Eqs.~\ref{eq:tc} and~\ref{eq:corex_expand}, we expand Eq.~\ref{eq:corex_lb} into basic information-theoretic quantities as follows: 
\begin{align}
\label{eq:corex_expand2}
\mathcal{L}(\theta;\vx)  &= TC_{\theta}(\vx;\vz) - TC_{\theta}(\vz)  \nonumber \\ 
&=\sum_{i=1}^{d} I_{\theta}(\vx_i:\vz) - I_\theta(\vx:\vz) - \sum_{i=1}^{m}  H_{\theta}(\vz_i) + H_\theta(\vz) \nonumber \\
&= \sum_{i=1}^{d} I_{\theta}(\vx_i:\vz) - \sum_{i=1}^{m}  H_{\theta}(\vz_i) + H_{\theta}(\vz|\vx) 
\end{align}
If we further constrain our search space $p_{\theta}(\vz|\vx)$ to have the factorized form $p_{\theta}(\vz|\vx) = \prod_{i=1}^{m} p_{\theta_i}(\vz_i|\vx)$\footnote{Each marginal distribution $p_{\theta_i}(\vz_i|\vx)$ is parametrized by a different $\theta_i$. But we will omit the subscript $i$ under $\theta$ for simplicity, as well as $\phi$, $\alpha$ in the following context.} which is a standard assumption in most VAE models, then we have:
\begin{equation} \label{eq:corex_mi}
\begin{split}
\mathcal{L}(\theta;\vx) &= TC_{\theta}(\vx;\vz) - TC_{\theta}(\vz)\\
&= \sum_{i=1}^{d} I_{\theta}(\vx_i:\vz) - \sum_{i=1}^{m}I_{\theta}(\vz_i:\vx)
\end{split}
\end{equation}
We convert the two total correlation terms into two sets of mutual information terms in Eq.~\ref{eq:corex_mi}. The first term, $I_{\theta}(\vx_i:\vz)$, denotes the mutual information between each input dimension $\vx_i$ and $\vz$, and can be broadly construed as measuring the ``relevance'' of the representation to each observed variable in the parlance of the information bottleneck~\cite{tishby2000information,shwartz2017opening}. 
The second term, $I_{\theta}(\vz_i:\vx)$, represents the mutual information between each latent dimension $\vz_i$ and $\vx$ and can be viewed as the compression achieved by each latent factor. We proceed by constructing tractable bounds on these quantities.

\subsection{Variational Lower Bound for $I_\theta(\vx_i:\vz)$}
~\cite{barber2003algorithm} derived the following lower bound for mutual information by using the non-negativity of KL-divergence; i.e., $\Sigma_{\vx_i}p(\vx_i|\vz)\log\frac{p(\vx_i|\vz)}{q(\vx_i|\vz)} \ge 0$ gives:
\begin{equation}\label{eq:xi_z}
I_\theta(\vx_i:\vz) \ge H(\vx_i) + \left<\ln q_{\phi}(\vx_i | \vz) \right>_{p_{\theta}(\vx, \vz)}
\end{equation}
where the angled brackets represent expectations, and $q_{\phi}(\vx_i | \vz)$ is any arbitrary distribution parametrized by $\phi$. We need a variational distribution $q_{\phi}(\vx_i | \vz)$ because the posterior distribution $p_\theta(\vx|\vz)=p_\theta(\vz|\vx)p(\vx)/p_\theta(\vz)$ is hard to calculate because the true data distribution $p(\vx)$ is unknown---although approximating the normalization factor $p_\theta(\vz)$ can be tractable compared to VAE. A detailed comparison with VAE will be made in Sec.~\ref{sec:connection}. 

\subsection{Variational Upper Bound for $I_\theta(\vz_i:\vx)$}

We again use the non-negativity of KL-divergence, i.e., $\Sigma_{\vz_i}p(\vz_i)\log\frac{p(\vz_i)}{r(\vz_i)} \ge 0$, to obtain:
\begin{align}
\nonumber
& I_\theta\left( {\vx:{\vz_i}} \right)  \\  \nonumber 
&= \int {d\vx d{\vz_i}p_\theta\left( {{\vz_{i,}}\vx} \right)\log p_\theta\left( {{\vz_i}|\vx} \right)}  - \int {d{\vz_i}p_\theta\left( {{\vz_i}} \right)\log p_\theta\left( {{\vz_i}} \right)}  \\  \nonumber 
&\le \int {d\vx d{\vz_i}p_\theta\left( {{\vz_{i,}}\vx} \right)\log p_\theta\left( {{\vz_i}|\vx} \right)}  - \int {d{\vz_i}p_\theta\left( {{\vz_i}} \right)\log r_\alpha\left( {{\vz_i}} \right)}  \\ \nonumber 
&= \int {d\vx d{\vz_i}p_\theta\left( \vx \right)p_\theta\left( {{\vz_i}|\vx} \right)\log \frac{{p_\theta\left( {{\vz_i}|\vx} \right)}}{{r_\alpha\left( {{\vz_i}} \right)}}}  \\  
&= D_{KL}\left( {p_\theta\left( {{\vz_i}|\vx} \right)||r_\alpha\left( {{\vz_i}} \right)} \right)  \label{eq:zi_x} 
\end{align}
where $r_\alpha(\vz_i)$ represents an arbitrary distribution parametrized by $\alpha$.

Combining bounds in Eqs.~\ref{eq:xi_z} and~\ref{eq:zi_x} into Eq.~\ref{eq:corex_mi}, we have:
\begin{equation} 
\label{eq:mi_lb}
\begin{split}
\mathcal{L}(\theta;\vx) = &\sum_{i=1}^{d} I_\theta(\vx_i:\vz) - \sum_{i=1}^{m}I_\theta(\vz_i:\vx)  \\
&\ge \sum_{i=1}^{d} H(\vx_i) + \left<\ln q_\phi(\vx_i | \vz) \right>_{p_\theta(\vx,\vz)} \\
- & \sum_{i=1}^{m}  D_{KL}(p_\theta(\vz_i|\vx)||r_\alpha(\vz_i))
\end{split}
\end{equation}
We then can jointly optimize the lower bound in Eq.~\ref{eq:mi_lb} w.r.t. both the stochastic parameter $\theta$ and the variational parameters $\phi$ and $\alpha$.

%% file: sections/connection.tex
\section{Connection to Variational Autoencoders} \label{sec:connection}
Remarkably, Eq.~\ref{eq:mi_lb} has a form that is very similar to the lower bound introduced in variational autoencoders, except it is decomposed into each dimension $\vx_i$ and $\vz_i$. To pursue this similarity further, we denote
\begin{equation}\label{eq:factor}
q_\phi(\vx|\vz) = \prod_{i=1}^{d}q_\phi(\vx_i|\vz), \quad 
r_\alpha(\vz) = \prod_{i=1}^{m}r_\alpha(\vz_i)
\end{equation}
Then, by rearranging the terms in Eq.~\ref{eq:mi_lb}, we obtain 
\begin{eqnarray}
\label{eq:auto_corex}
\mathcal{L}(\theta;\vx)  &=& \sum_{i=1}^{d} I_\theta(\vx_i:\vz) - \sum_{i=1}^{m}I_\theta(\vz_i:\vx) \nonumber \\
&\ge& 
\left(\sum_{i=1}^{d} H(\vx_i)\right)  + \left<\ln \underbrace{q_\phi(\vx | \vz)}_{\text{decoder}} \right>_{p_\theta(\vx,\vz)} \nonumber \\ &-& D_{KL}(\underbrace{p_\theta(\vz|\vx)}_{\text{encoder}}||r_\alpha(\vz))
\end{eqnarray}
The first term in the bound, $\sum_{i=1}^{d} H(\vx_i)$, is a constant and has no effect on the optimization. The remaining expression coincides with the VAE objective as long as $r_\alpha(\vz)$ is a standard Gaussian. The second term corresponds to the reconstruction error, and  the third term is the KL-divergence term in VAE.  

\paragraph{Comparison} The CorEx objective starts with a defined encoder $p_\theta(\vz|\vx)$ and seeks a decoder $q_\phi(\vx|\vz)$ via variational approximation to the true posterior. VAE is exactly the opposite. Moreover, in VAE we need a variational approximation to the posterior because the normalization constant is intractable; in CorEx the variational distribution is needed because we do not know the true data distribution $p(\vx)$. It is also worth mentioning that the lower bound in Eq.~\ref{eq:auto_corex} requires a fully factorized form of the decoder $q_\phi(\vx|\vz)$, unlike VAE where $q_\phi(\vx|\vz)$ can be flexible.\footnote{In this paper we also restrict the encoder distribution $p_\theta(\vz|\vx)$ to have a factorized form which follows the standard network structures in VAE, but it is not a necessary condition to achieve the lower bound shown in Eq.~\ref{eq:auto_corex}. } 

As pointed out by~\cite{zhao2017infovae}, if we choose to use a more expressive distribution family, such as PixelRNN/PixelCNN~\cite{van2016pixel, gulrajani2016pixelvae} for the decoder in a VAE, the model tends to neglect the latent codes altogether, i.e., $I(\vx:\vz)=0$. This problem, however, does not exist in CorEx, since it explicitly requires $\vz$ to be \textit{informative} about $\vx$ in the objective function. It is this \textit{informativeness} term that leads the CorEx objective to a factorized decoder family $q_\phi(\vx|\vz)$. In fact, if we assume $I_\theta(\vx:\vz) = 0$, then we will get $TC(\vx)=TC_\theta(\vx|\vz)$ and an \textit{informativeness} term $TC_\theta(\vx;\vz)$ of zero---meaning CorEx will avoid such undesirable solutions. 

\paragraph{Stacking CorEx and Hierarchical VAE}\comment{\aram{Is this supposed to be a section?}} Notice that if Eq.~\ref{eq:corex_lb} does not achieve the global maximum, it might be the case that the latent variable $\vz$ is still not disentangled enough, i.e., $TC_{\theta}(\vz) > 0$. If this is true, we can reapply the CorEx principle~\cite{ver2015maximally} and learn another layer of latent variables $\vz^{(2)}$ on top of $\vz$ and redo the optimization on $\theta^{(2)}$ w.r.t. the following equation; i.e.,
\begin{align} \label{eq:corex_lb_1}
\mathcal{L}(\theta^{(2)};\vz) &= TC_{\theta^{(2)}}(\vz;\vz^{(2)}) - TC_{\theta^{(2)}}(\vz^{(2)}) \\
&= TC_{\theta}(\vz) - TC_{\theta^{(2)}}(\vz|\vz^{(2)}) - TC_{\theta^{(2)}}(\vz^{(2)}) \nonumber
\end{align}
To generalize, suppose there are $L$ layers of latent variables, $\vz^{(1)}, \vz^{(2)}, ..., \vz^{(L)}$ and we further denote the observed variable $\vx\equiv\vz^{(0)}$.  Then one can stack each latent variable $\vz^{(l)}$ on top of $\vz^{(l-1)}$  and jointly optimize the summation of the corresponding objectives, as shown in Eqs.~\ref{eq:corex_lb} and~\ref{eq:corex_lb_1}; i.e., 
\begin{equation} \label{eq:corex_lb_stack}
\mathcal{L}(\theta^{(1,2,..,L)};\vx) = \sum_{l=1}^{L} \mathcal{L}(\theta^{(l)}; \vz^{(l-1)})
\end{equation}
By simple expansion of Eq.~\ref{eq:corex_lb_stack} and cancellation of intermediate $TC$ terms, we have:
\begin{align} 
&\mathcal{L}(\theta^{(1,2,..,L)};\vx) \nonumber \\
&=\mathcal{L}(\theta^{(1)};\vz^{(0)})+\mathcal{L}(\theta^{(2)};\vz^{(1)})+...+\mathcal{L}(\theta^{(L)};\vz^{(L-1)}) \nonumber \\
& = TC(\vx) - \sum_{l=1}^{L} TC_{\theta^{(l)}}(\vz^{(l-1)} | \vz^{(l)}) - TC_{\theta^{(L)}}(\vz^{(L)}) \nonumber \\
& \le TC(\vx) \label{eq:corex_lb_stack_lb} 
\end{align}
Furthermore, if we have $\mathcal{L}(\theta^{(l)};\vz^{(l-1)}) > 0$ for all $l$, then we get:
\begin{align} 
\begin{split}
\mathcal{L}(\theta^{(1)};\vx) \le \mathcal{L}(\theta^{(1,2)};\vx) \le ... &\le \mathcal{L}(\theta^{(1,...,L)};\vx) \\
&\le TC(\vx)
\end{split}
\label{eq:corex_lb_stack_prog} 
\end{align}

Eq.~\ref{eq:corex_lb_stack_prog} shows that stacking latent factor representations results in progressively better lower bounds for $TC(\vx)$. 

To optimize Eq.~\ref{eq:corex_lb_stack}, we reuse Eqs.~\ref{eq:corex_mi},~\ref{eq:xi_z} and ~\ref{eq:zi_x} and get: 
\begin{eqnarray}
\mathcal{L}(\theta^{(1,2,..,L)};\vx) &\ge& \sum_{i} H(\vz^{(0)}_i)  \nonumber \\ 
&+& \sum_{l=1}^{L}\sum_{i} \left<\ln q_{\phi^{(l)}}(\vz^{(l-1)}_i | \vz^{(l)}) \right>_{p_\theta(\vz)} \nonumber \\
&-&  \sum_{l=1}^{L} \sum_{i}  \left<\ln p_{\theta^{(l)}}(\vz^{(l)}_i | \vz^{(l-1)}) \right>_{p_\theta(\vz)}  \nonumber \\
 &+& \sum_{i} \left<\ln r_{\alpha}(\vz_i^{(L)})\right>_{p_\theta(\vz)} 
\end{eqnarray}
Enforcing independence relations at each layer, we denote:
\be\label{eq:factor_stack}
q_\phi(\vx, \vz) &=\prod_i r_{\alpha}(\vz_i^{(L)})\times \prod_{l=1}^{L}\prod_{i}q_{\theta^{(l)}}(\vz_i^{(l-1)}|\vz^{(l)})\\ 
p_\theta(\vz|\vx)&=\prod_{l=1}^{L} p_{\theta^{(l)}}(\vz^{(l)}|\vz^{(l-1)}) \\
\ee
and obtain
\begin{eqnarray}\label{eq:auto_corex_stack}
\mathcal{L}(\theta^{(1,2,..,L)};\vx) &\ge& \sum_{i} H(\vz^{(0)}_i)  \nonumber \\
 &+& \left<\ln \frac{q_\phi(\vx,\vz)}{p_\theta(\vz|\vx)}\right>_{p_\theta(\vz|\vx)p(\vx)}
\end{eqnarray}
One can now see that the second term of the RHS in Eq.~\ref{eq:auto_corex_stack} has the same form as deep latent Gaussian models~\cite{rezende2014stochastic} (also known as hierarchical VAE) as long as the latent code distribution $r_\alpha(\vz^{(L)})$ on the top layer follows standard normal and $q_{\theta^{(l)}}(\vz^{(l-1)}|\vz^{(l)})$ on each layer is parametrized by Gaussian distributions.

One immediate insight from this connection is that, as long as each $\mathcal{L}(\theta^{(l)}; \vz^{(l-1)}$) is greater than zero in Eq.~\ref{eq:corex_lb_stack},  then by expanding the definition of each term we can easily see that $\vz^{(l)}$ is more disentangled than $\vz^{(l-1)}$; i.e., $TC(\vz^{(l-1)}) > TC(\vz^{(l)})$ if $TC(\vz^{(l-1)})-TC(\vz^{(l-1)}|\vz^{(l)})-TC(\vz^{(l)}) > 0$. Therefore, each latent layer of hierarchical VAE will be more and more disentangled if $\mathcal{L}(\theta^{(l)}; \vz^{(l-1)}) > 0$ for each $l$. This interpretation also provides a criterion for determining the depth of a hierarchical representation; we can add layers as long as the corresponding term in the objective is positive so that the overall lower bound on $TC(\vx)$ is increasing. 

Despite reaching the same final expression, approaching this result from an information-theoretic optimization rather than generative modeling perspective offers some advantages. First, we have much more flexibility in specifying the distribution of latent factors, as we can directly sample from this distribution using our encoder. Second, the connection with mutual information suggests intuitive modifications of our objective that increase the interpretability of results. These advantages will be explored in more depth in Sec.~\ref{sec:app}.


%% file: sections/results.tex
\section{Applications}\label{sec:app}

\subsection{Disentangling Latent Codes via Hierarchical VAE / Stacking CorEx on MNIST}\label{sec:app_mnist}
\begin{figure}[!h]
	\centering
	\includegraphics[width=0.7\columnwidth]{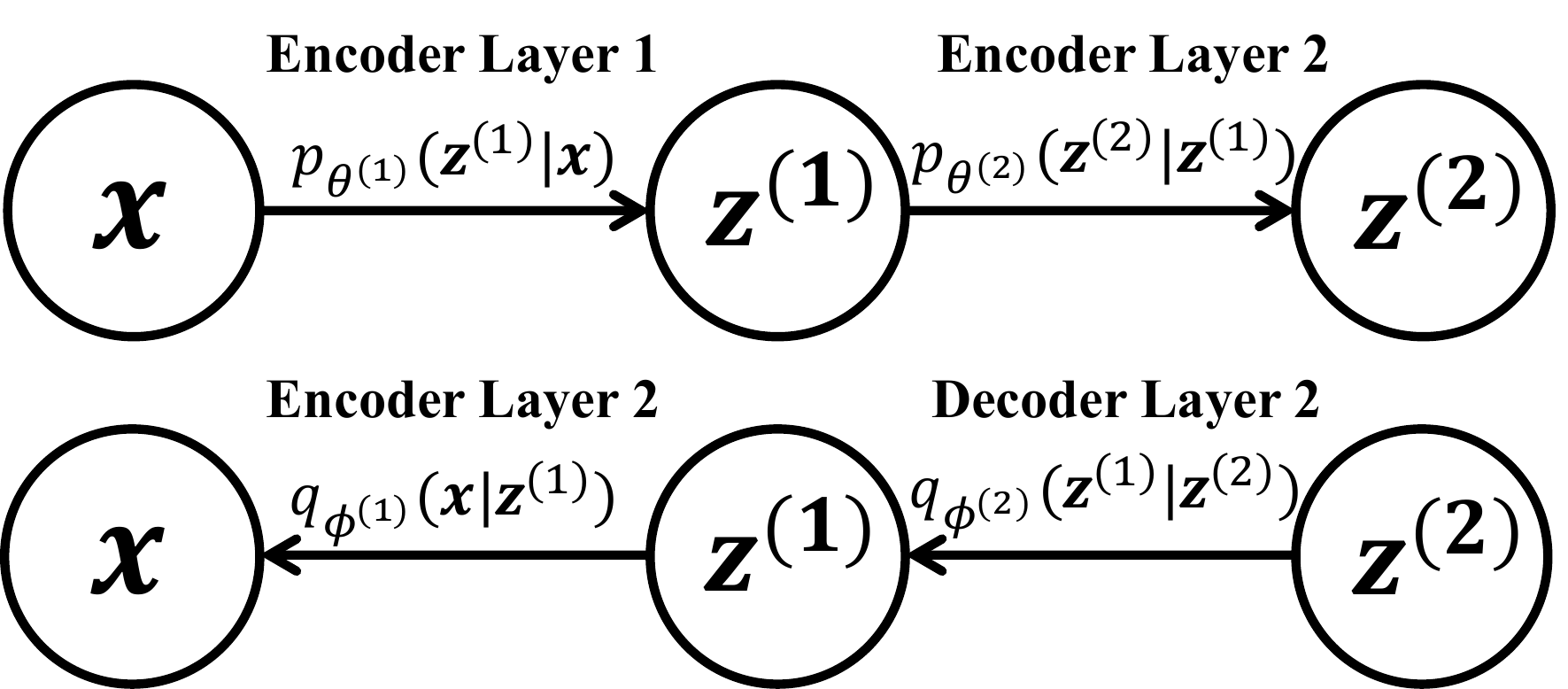}
	\caption{Encoder and decoder models for MNIST, where $\vz^{(1)}$ is 64 dimensional continuous variable and $\vz^{(2)}$ is a discrete variable (one hot vector with length ten). }
	\label{fig:mnist_g}
\end{figure}

We train a simple hierarchical VAE/stacking CorEx model with two stochastic layers on the MNIST dataset.  The graphical model is shown in Fig.~\ref{fig:mnist_g}. For each stochastic layer, we use a neural network to parametrize the distribution $p_\theta$ and $q_\phi$, and we set $r_\alpha$ to be a fixed standard Gaussian. 
We use a 784-512-512-64 fully connected network between $\vx$ and $\vz^{(1)}$ and a 64-32-32-16-16-10 dense network between $\vz^{(1)}$ and $\vz^{(2)}$, with ReLU activations in both. The output of $\vz^{(2)}$ is a ten-dimensional one hot vector, where we decode based on each one-hot representation and weight the results according to their softmax probabilities.

After training the model, we find that the learned discrete variable $\vz^{(2)}$ on the top layer gives us an unsupervised classification accuracy of 85\%, which is competitive with the more complex method shown in~\cite{dilokthanakul2016deep}. 

\begin{figure}[ht]
	\centering
\includegraphics[width=0.82\columnwidth]{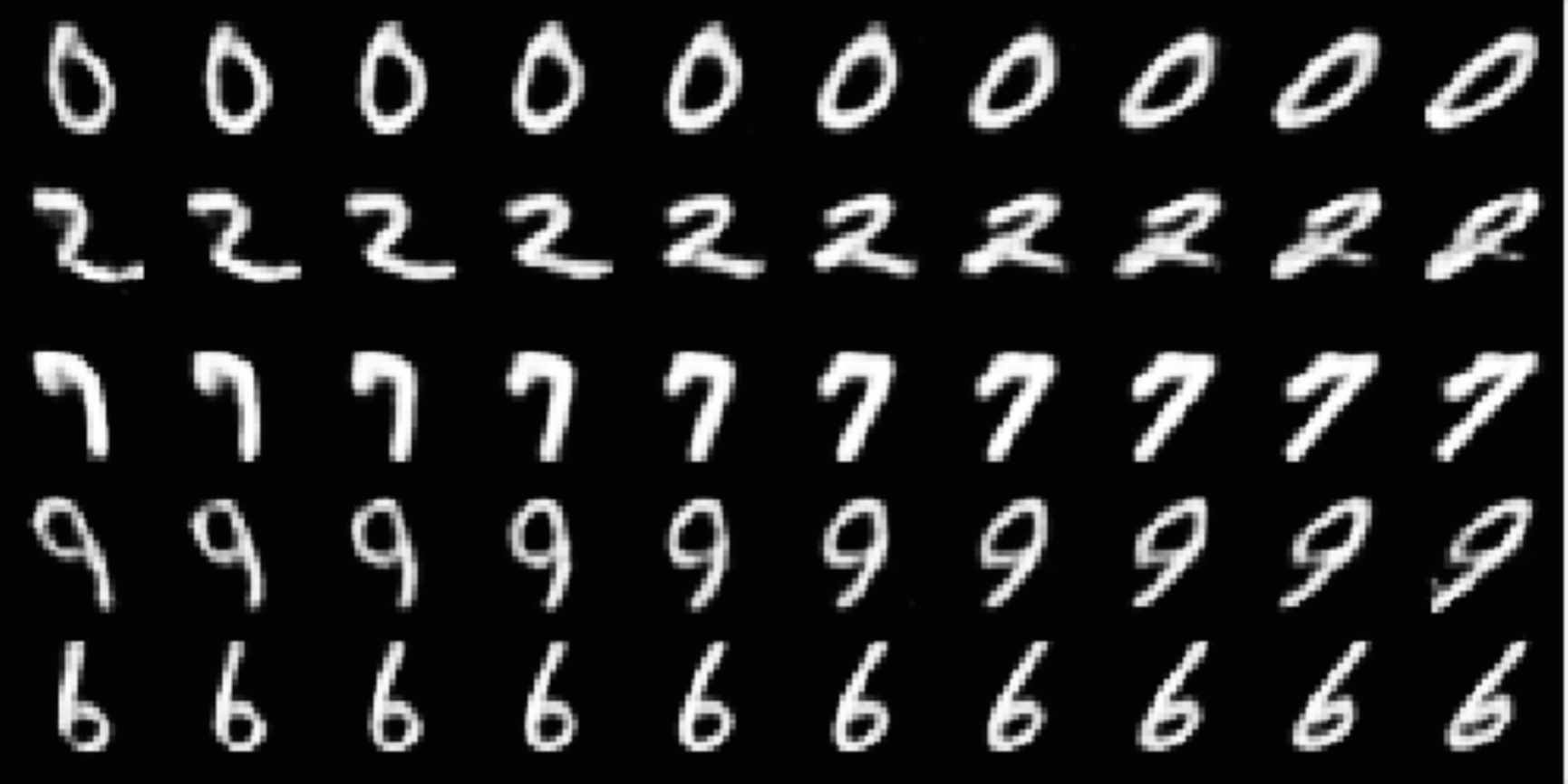} \\
		{(a) Manipulating $\vz^{(1)}_a$ with MNIST. (Azimuth)} \\
\includegraphics[width=0.82\columnwidth]{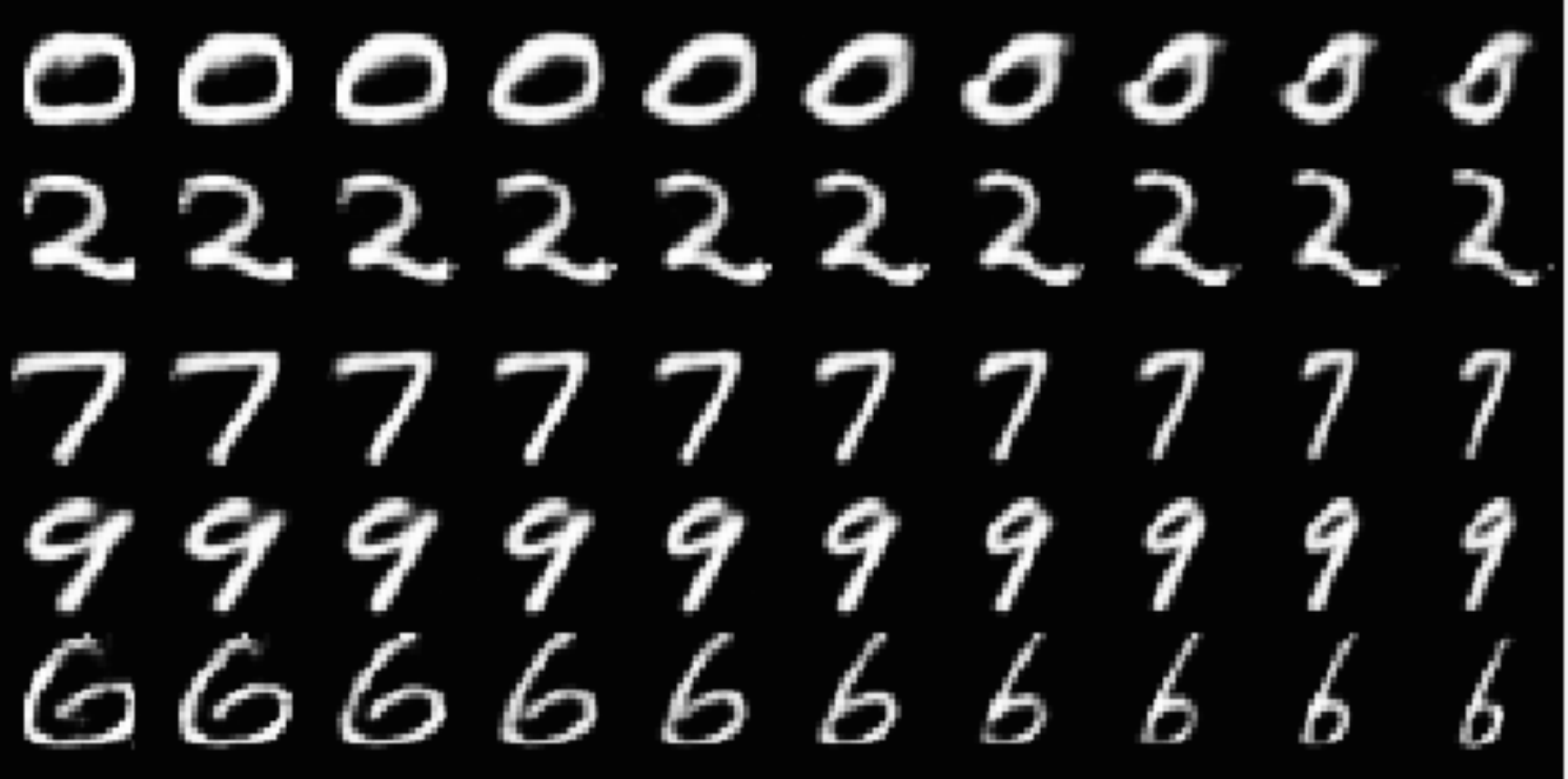} \\
		{(b) Manipulating $\vz^{(1)}_b$ with MNIST. (Width)}
		\caption{\textbf{Varying the latent codes of $\vz^{(1)}$ on MNIST:} In both figures, each row corresponds to a fixed discrete number in layer $\vz^{(2)}$. Different columns correspond to the varying noise from the selected latent node in layer $\vz^{(1)}$ from left to right, while keeping other latent codes fixed. In (a) varying the noise results in different rotations of the digit; In (b) a small (large) value of the latent code corresponds to wider (narrower) digit.}
		\label{fig:mnist_int}
\end{figure}

To verify that the top layer $\vz^{(2)}$ helps disentangle the middle layer $\vz^{(1)}$ by encouraging conditional independence of $\vz^{(1)}$ given $\vz^{(2)}$, we calculate the mutual information $I_\theta(\vx:\vz^{(1)}_i)$ between input $\vx$ and each dimension $\vz^{(1)}_i$. \comment{, as shown in Fig.~\ref{fig:mnist_mi}. We can see around 80\% of the latent codes have very low mutual information with $\vx$.} We then select the top two dimensions with the most mutual information, and denote these two dimensions as $\vz^{(1)}_a$, $\vz^{(1)}_b$. We find $I_\theta(\vx:\vz^{(1)}_a)=2.71$ and $I_\theta(\vx:\vz^{(1)}_b)=2.56$. We then generate new digits by first fixing the discrete latent variable $\vz^{(2)}$ on the top layer, and sampling latent codes $\vz^{(1)}$ from $q_\phi(\vz^{(1)}|\vz^{(2)})$.  We systematically vary the noise from -2 to 2 through $q_\phi(\vz^{(1)}_a|\vz^{(2)})$ and $q_\phi(\vz^{(1)}_b|\vz^{(2)})$ while keeping the other dimensions of $\vz^{(1)}$ fixed, and visualize the results in Fig.~\ref{fig:mnist_int}.
We can see that this simple two-layer structure automatically disentangles and learns the interpretable factors on MNIST (width and rotation). We attribute this behavior to stacking, where the top layer disentangles the middle layer and makes the latent codes more interpretable through samples from $q_\phi(\vz^{(1)}|\vz^{(2)})$.
\subsection{Learning Interpretable Representations through Information Maximizing VAE / CorEx on CelebA }
One important insight from recently developed methods, like InfoGAN, is that we can maximize the mutual information between a latent code and the observations to make the latent code more interpretable. 

While it seems ad hoc to add an additional mutual information term in the original VAE objective, a more natural analogue arises in the CorEx setting. Looking at the formulation in Eq.~\ref{eq:corex_mi}, it already contains two sets of mutual information terms. If one would like to {\em anchor} a latent variable, say $\vz_a$, to have higher mutual information with the observation $\vx$, then one can simply modify the objective by replacing the unweighted sum with a weighted one: 
\begin{align} 
\label{eq:corex_anchored}
&\mathcal{L}_{anchor}(\theta;\vx) \\ \nonumber
&= TC_{\theta}(\vx;\vz) - TC_{\theta}(\vz) + \lambda I_\theta(\vz_a:\vx)\\ \nonumber
&= \sum_{i=1}^{d} I_{\theta}(\vx_i:\vz) - \sum_{i=1, i \neq a}^{m}I_{\theta}(\vz_i:\vx) - (1-\lambda) I_\theta(\vz_a:\vx) \nonumber 
\end{align}
Eq.~\ref{eq:corex_anchored} suggests that mutual information maximization in CorEx is achieved by modifying the corresponding weights of the second term $I_\theta(\vz_i:\vx)$ in Eq.~\ref{eq:corex_mi}. We then use the lower bound in Eq.~\ref{eq:mi_lb} to obtain
\begin{eqnarray} \label{eq:vae_anchored}
\mathcal{L}_{anchor}(\theta;\vx) &\ge& \sum_{i=1}^{d} H(\vx_i)  \\
&+& \left<\ln q_\phi(\vx_i | \vz) \right>_{p_\theta(\vx,\vz)} \nonumber \\
&-&  \sum_{i=1, i \neq a}^{m}  D_{KL}(p_\theta(\vz_i|\vx)||r_\alpha(\vz_i)) \nonumber \\
&-& (1-\lambda)D_{KL}(p_\theta(\vz_a|\vx)||r_\alpha(\vz_a)) \nonumber
\end{eqnarray}

Eq.~\ref{eq:vae_anchored} shows that in VAE we can decrease the weight of KL-divergence for particular latent codes to achieve mutual information maximization. We call this new approach \textbf{AnchorVAE} in Eq.~\ref{eq:vae_anchored}. Notice that there is a subtle difference between AnchorVAE and $\beta$-VAE~\cite{higgins2016beta}. In $\beta$-VAE, the weights of KL-divergence term for all latent codes are the same, while in AnchorVAE, only the weights of specified factors have been changed to encourage high mutual information.  With some prior knowledge of the underlying factors of variation, AnchorVAE encourages the model to concentrate this explanatory power in a limited number of variables.
\begin{figure}[htbp]
	\centering
\includegraphics[width=0.7\columnwidth]{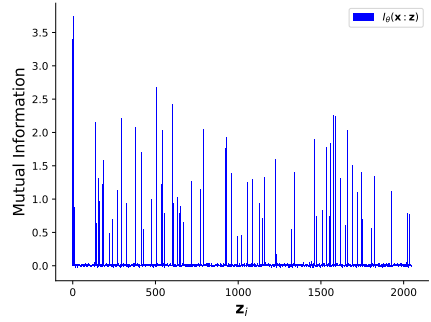}
	\caption{Mutual information between input data $\vx$ and each latent variable $\vz_i$ in CelebA with AnchorVAE. It is clear that the anchored first five dimensions have the highest mutual information with $\vx$.}
	\label{fig:celeba_mi}
\end{figure}

We trained AnchorVAE on the CelebA dataset with 2048 latent factors, with mean square error for reconstruction loss. We adopted a three-layer convolutional neural network structure. The weights of KL-divergence of the first five latent variables are set to 0.5 to let them have higher mutual information than other latent variables. The mutual information is plotted in Fig.~\ref{fig:celeba_mi} after training. We find these five latent variables have the highest mutual information of around 3.5, demonstrating the mutual information maximization effect in AnchorVAE.

\begin{figure}[!h]
\centering
\begin{subfigure}[t]{0.49\columnwidth}
\centering
\includegraphics[width=1.0\columnwidth]{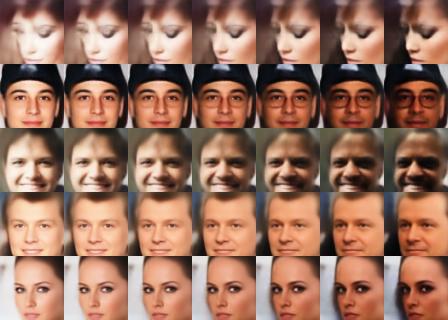} 
\caption{Varying $\vz_0$. (Skin Color)} 
\end{subfigure}
\begin{subfigure}[t]{0.49\columnwidth}
\centering
\includegraphics[width=1.0\columnwidth]{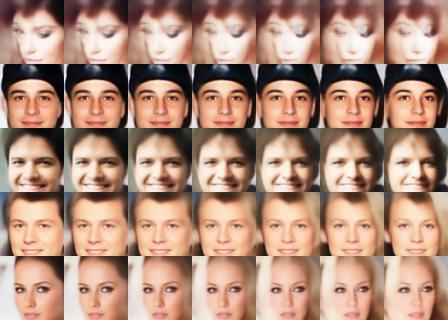} 
\caption{Varying $\vz_1$. (Azimuth)}
\end{subfigure}

\begin{subfigure}[t]{0.49\columnwidth}
\centering
\includegraphics[width=1.0\columnwidth]
{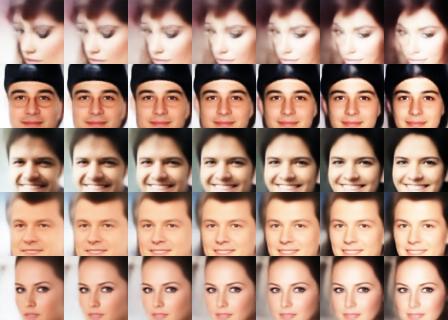}
\caption{Varying $\vz_2$. (Emotion)}
\end{subfigure}
\begin{subfigure}[t]{0.49\columnwidth}
\centering
\includegraphics[width=1.0\columnwidth]{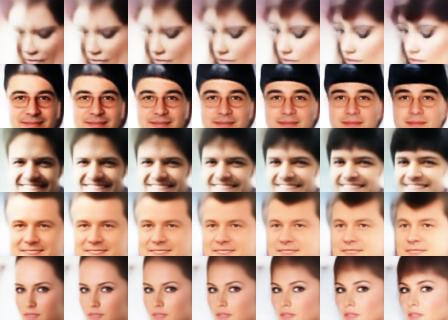}
\caption{Varying $\vz_3$. (Hair)}
\end{subfigure}
\begin{subfigure}[!h]{0.5\columnwidth}
\centering
\includegraphics[width=1.0\columnwidth]
{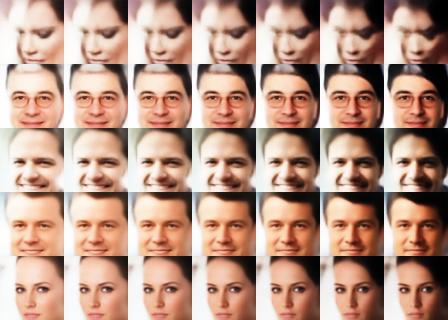}
\caption{Varying $\vz_4$. (Lighting)}
\end{subfigure}
\caption{\textbf{Manipulating latent codes $\vz_0, \vz_1, \vz_2, \vz_3, \vz_4$ on CelebA using AnchorVAE}: We show the effect of the anchored latent variables on the outputs while traversing their values from [-3,3]. Each row represents a different seed image to encode latent codes. Each anchored latent code represents a different factor on interpretablility. (a) Skin Color (b) Azimuth (c) Emotion (Smile) (d) Hair (less or more) (e) Lighting.}
\label{fig:celeba_int}
\end{figure}

To evaluate the interpretability of those anchored variables for generating new samples, we manipulate the first five latent variables while keeping other dimensions fixed. Fig.~\ref{fig:celeba_int} summarizes the result. We observe that all five anchored latent variables learn intuitive factors of variation in the data. It is interesting to see that latent variable $\vz_0$ and $\vz_4$ are very similar---both vary the generated images from white to black in some sense. However, these two latent factors are actually very different: $\vz_0$ emphasizes skin color variation while $\vz_4$ controls the position of the light source. 

\begin{figure}[!h]
\centering
\begin{subfigure}[t]{0.49\columnwidth}
\centering
\includegraphics[width=1.0\columnwidth]{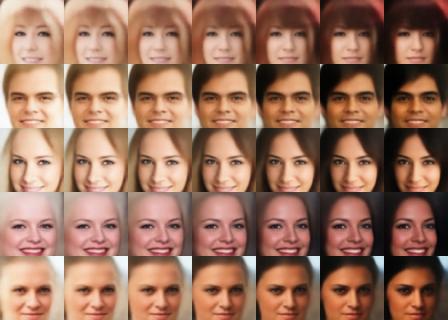} 
\caption{$\vz_{130}$ entangles skin color with hair} 
\end{subfigure}
\begin{subfigure}[t]{0.49\columnwidth}
\centering
\includegraphics[width=1.0\columnwidth]{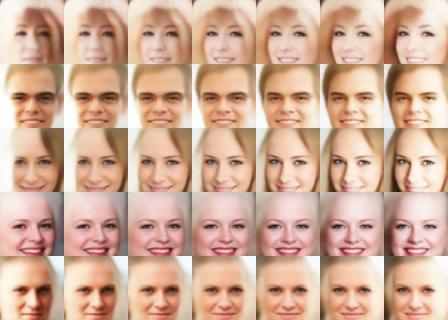} 
\caption{$\vz_{610}$ entangles emotion with azimuth}
\end{subfigure}
\caption{\textbf{Manipulating top two latent codes with the most mutual information on CelebA using original VAE}. We observe that both latent codes learned entangled representations. (a) $\vz_{130}$ entangles skin color with hair; (b) $\vz_{610}$ entangles emotion with azimuth.}
\label{fig:celeba_naive}
\end{figure}

We also trained the original VAE objective with the same network structure and examine the top five latent codes with highest mutual information. Fig.~\ref{fig:celeba_naive} shows the results of manipulating the top two latent codes $\vz_{130}$, $\vz_{610}, $ with mutual information $I(\vz_{130}:\vx)=3.1$ and $I(\vz_{610}:\vx)=2.8$ respectively. We can see that they reflect an entangled representation. The other three latent codes demonstrate similar entanglements which are omitted here.
\subsection{Generating Richer and More Realistic Images via CorEx}
Let us revisit the variational upper bound on $I_\theta(\vx:\vz_i)$ in Eq.~\ref{eq:zi_x}. In this upper bound, VAE chooses $r_\alpha\left( {{\vz_i}} \right)$ to be a standard normal distribution. But notice that this upper bound becomes tight when $r_\alpha\left( {{\vz_i}} \right) = p_\theta(\vz_i)$; i.e., 
\be
 I_\theta\left( {\vx:{\vz_i}} \right) \equiv  D_{KL}\left( {p_\theta\left( {{\vz_i}|\vx} \right)||p_\theta\left( {{\vz_i}} \right)} \right) \\
\le D_{KL}\left( {p_\theta\left( {{\vz_i}|\vx} \right)||r_\alpha\left( {{\vz_i}} \right)} \right)
 \nonumber \\
\ee

where $p_\theta(\vz_i) = \int_\vx p_\theta(\vz_i|\vx) p(\vx) d\vx$. Therefore, after training the model, we can approximate the true distribution $p_\theta(\vz_i) \approx \frac{1}{N}\sum_{i=1}^{N} p_\theta(\vz_i|\vx^{[i]}) $ by first sampling a data point $\vx^{[i]}$ and then sampling from the conditional $p_\theta(\vz_i|\vx^{[i]})$. Repeating this process across latent dimensions, we can use the factorized distribution $\prod_{i=1}^{m} p_\theta(\vz_i)$ to generate new data instead of sampling from a standard normal. In this way, we obtain more realistic images since we are sampling from a tighter lower bound to the CorEx objective. 

\begin{figure}[ht]
\centering
\begin{subfigure}[t]{0.49\columnwidth}
\centering
\includegraphics[width=1.0\columnwidth]{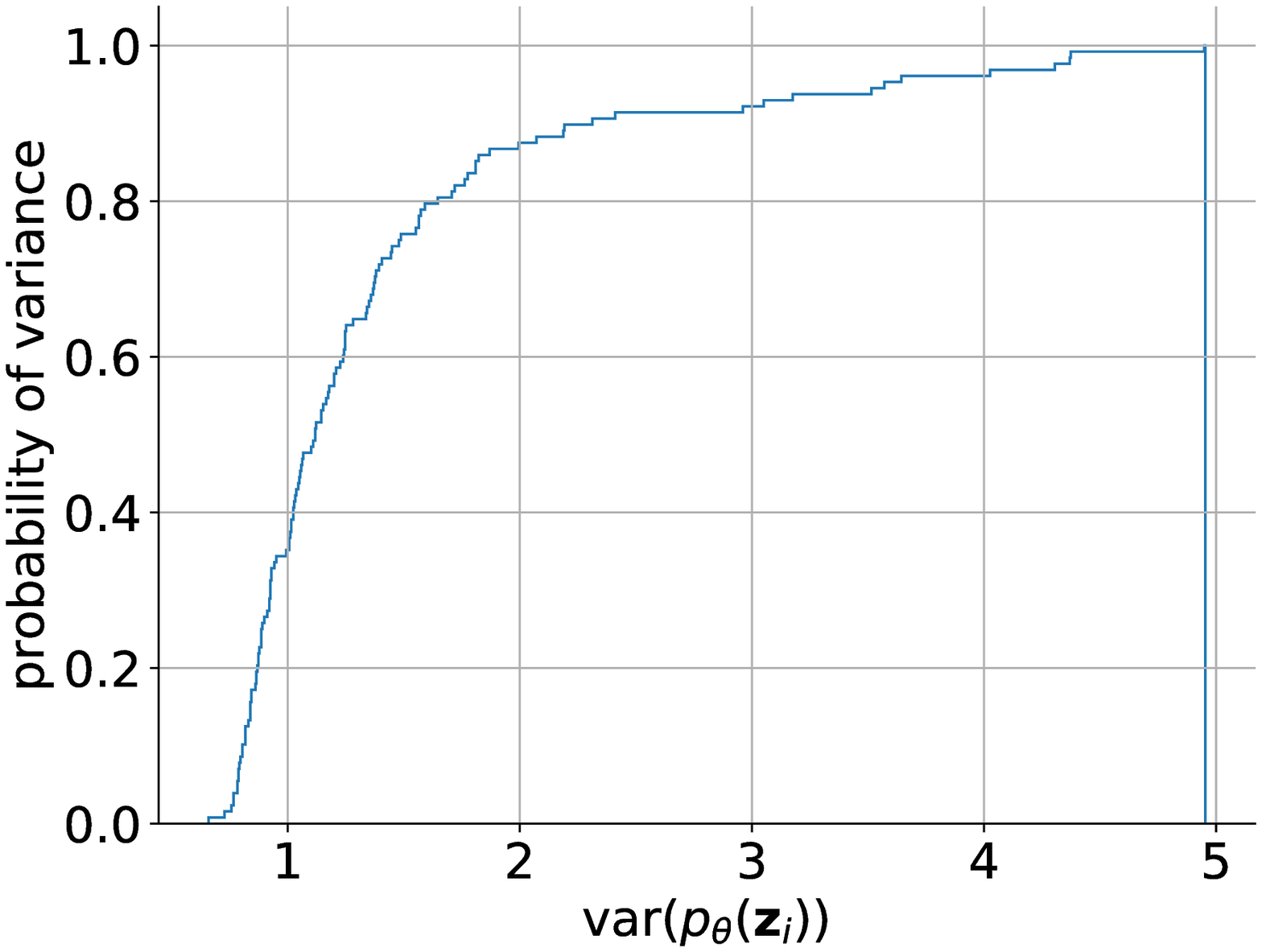} 
\caption{Cumulative distribution of variance for each $p_\theta(\vz_i)$)} 
\label{fig:celeba_var}
\end{subfigure}
\begin{subfigure}[t]{0.49\columnwidth}
\centering
\includegraphics[width=1.0\columnwidth]{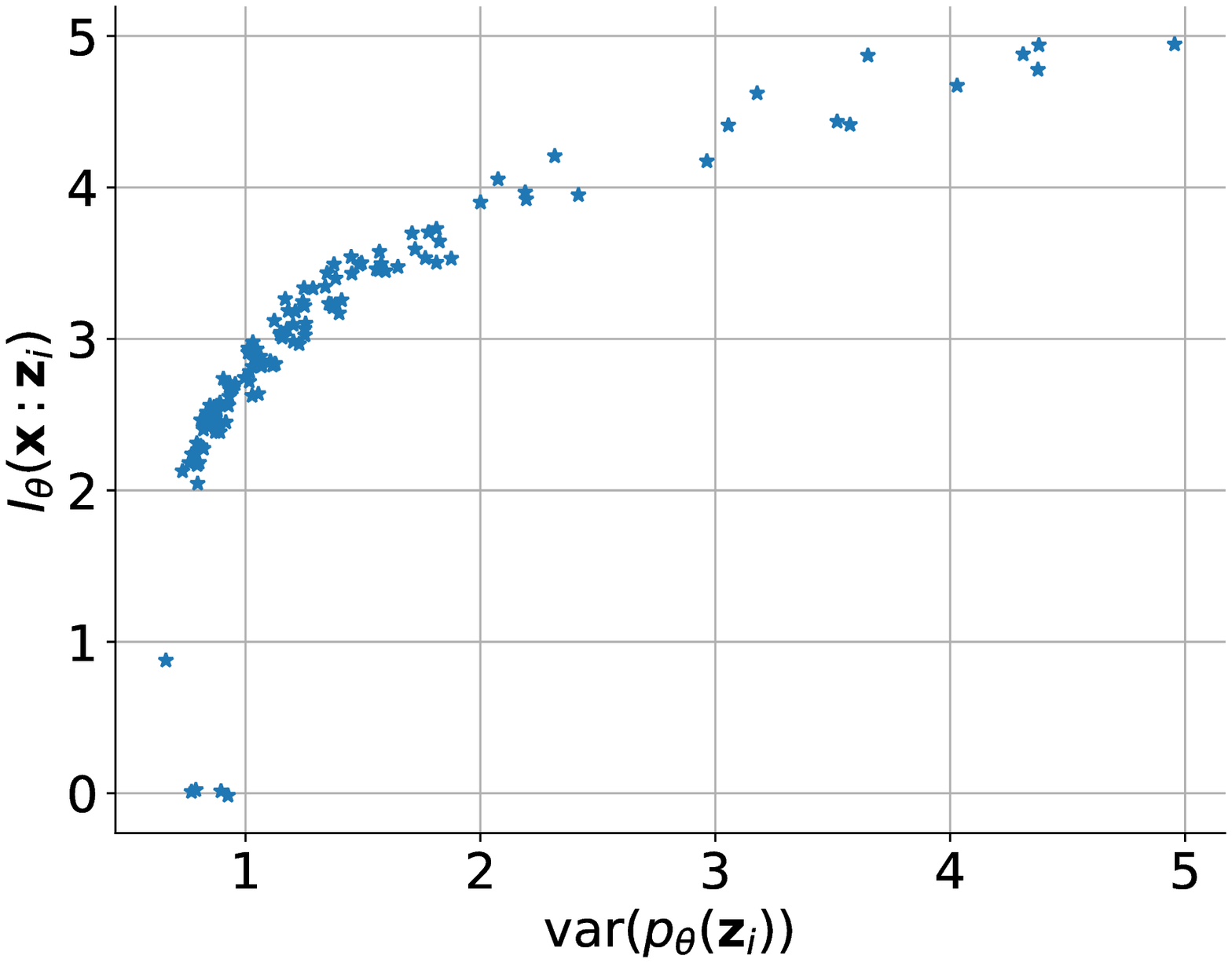} 
\caption{Variance of $p_\theta(\vz_i)$ versus mutual information $I_\theta(\vx:\vz_i)$} 
\label{fig:celeba_var_mi}
\end{subfigure}
\caption{Variance statistics for $p_\theta(\vz)$ on celebA after training a standard VAE with 128 latent codes.}
\label{fig:celeba_var_stat}
\end{figure}

We ran a traditional VAE on the celebA dataset with the log-normal loss as the reconstruction error and 128 latent codes. We calculated the variance of each $p_\theta(\vz_i)$ and ploted the cumulative distribution of these variances in Fig.~\ref{fig:celeba_var}.
One can see that around 20\% of the latent variables actually have a variance greater than two. We have plotted variance versus the mutual information in Fig.~\ref{fig:celeba_var_mi}, in which we can see that higher variance in $\vz_i$ corresponds to higher mutual information $I(\vx:\vz_i)$. In this case, using a standard normal distribution with variance 1 for all $\vz_i$ would be far from optimal for generating the data. 
\begin{figure}[htbp]
\centering
\begin{subfigure}[htbp]{0.49\columnwidth}
\centering
\includegraphics[width=1.0\columnwidth]{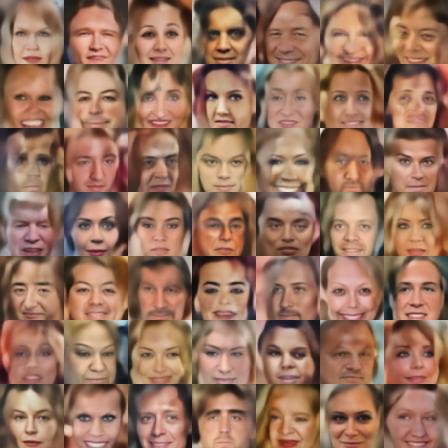} 
\caption{Latent codes are generated from standard normal} 
\label{fig:celeba_std}
\end{subfigure}
\begin{subfigure}[htbp]{0.49\columnwidth}
\centering
\includegraphics[width=1.0\columnwidth]{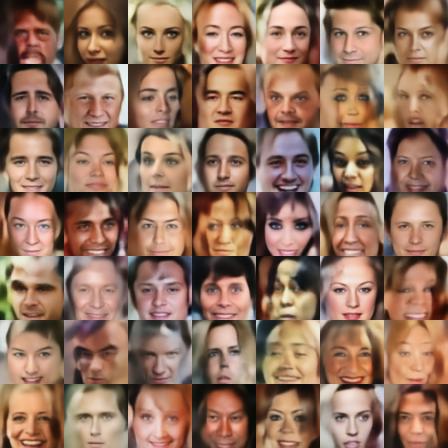} 
\caption{Latent codes are generated from $\prod_{i=1}^{m} p_\theta(\vz_i)$} 
\label{fig:celeba_margin}
\end{subfigure}
\caption{\textbf{Different sampling strategies of latent codes for CelebA dataset on VAE / CorEx.} Sampling latent codes from $\prod_{i=1}^{m} p_\theta(\vz_i)$ in (b) yields better quality images than sampling from a standard normal distribution in (a).}
\label{fig:celeba_gen}
\end{figure}

Fig.~\ref{fig:celeba_gen} shows the generated images by either sampling the latent code from a standard normal distribution or the factorized distribution $\prod_{i=1}^{m} p_\theta(\vz_i)$. We can see that Fig.~\ref{fig:celeba_margin} not only tends to generate more realistic images than Fig.~\ref{fig:celeba_std}, but  it also exhibits more diversity than Fig.~\ref{fig:celeba_std}. We attribute this improvement to the more flexible nature of our latent code distribution.

%% file: sections/conclusion-related.tex
\section{Related Work}\label{sec:related}
The notion of disentanglement in representation learning lacks a unique characterization, but it generally refers to latent factors which are individually interpretable, amenable to simple downstream modeling or transfer learning, and invariant to nuisance variation in the data \cite{bengio2013representation}.  We adopt the common definition of statistical independence \cite{achille2017emergence, nice} by minimizing total correlation---an idea with a rich history~\cite{barlow,comon,factorial}. However, there are numerous alternatives not rooted in independence.  \cite{higgins2016beta} measures disentanglement by the identifiability of changes in a single latent dimension.  More concretely, they vary only one latent variable with others fixed, apply the learned decoder and encoder to reconstruct the latent space, and propose that a classifier should be able to predict the varied dimension for a disentangled representation.  The work of \cite{thomas2017independently,bengio2017independently} is similar in spirit, identifying disentangled factors as changes in a latent embedding that can be controlled via reinforcement learning.  Alternatively, if prior knowledge of the number of desired factors of variation is given, models such as InfoGAN \cite{chen2016infogan} or our AnchorVAE seek to directly incorporate this information.

Our work provides a complementary perspective to a growing body of research connecting information theory and variational inference \cite{achille2017emergence,achille2018information,alemi2016deep}; much of this is motivated by the Information Bottleneck (IB) method \cite{tishby2000information}.  In the unsupervised case, IB generalizes the VAE objective by adding a Lagrange multiplier $\beta$ to the KL divergence term of the ELBO to manage the trade-off between data reconstruction and model compression.  This is identical to the $\beta$-VAE objective, where \cite{higgins2016beta} observes that overweighting the KL divergence term ($\beta > 1$) can encourage disentanglement, albeit at the cost of reconstruction performance.  \cite{achille2018information} add additional total correlation regularization to the IB Lagrangian to encourage independence, and propose using $\beta < 1$ and increasing $\beta$ gradually during training.  Furthermore, their optimization using multiplicative noise generalizes dropout methods, which helps to achieve improved robustness to nuisance variables.  

These objectives match CorEx and the ELBO for $\beta = 1$, but adding a Lagrange multiplier to control the disentangling term $TC(\vz)$ in CorEx would not lead to $\beta$-VAE. We saw in Sec. \ref{sec:connection} that our bound on the CorEx objective reduces to the ELBO with common factorization assumptions, so adding $\beta$ to $TC(\vz)$ in CorEx would lead to $TC(\vx;\vz) - \beta TC(\vz) \equiv  TC(\vx;\vz) - TC(\vz) - (\beta-1)TC(\vz) \ge ELBO - (\beta - 1) TC(\vz)$.  This bound recovers the objective of~\cite{kimdisentangling}, who consider $\beta > 1$ to encourage independence, but without the more principled justification of CorEx.

\cite{sonderby2016ladder, zhao2017learning} highlight limitations of the naive hierarchical VAE, such as representational inefficiency, and propose alternative ladder neural network structures for learning hierarchical features. However, from the CorEx perspective, we observe that the hierarchical VAE is encouraging more disentangled representations in top layers, which has not been previously recognized. 


\section{Conclusion}\label{sec:con}

Deep learning enables us to construct latent representations that reconstruct or generate samples from complex, high-dimensional distributions. Unfortunately, these powerful models do not necessarily produce representations with structures that match human intuition or goals. Subtle changes to training objectives lead to qualitatively different representations, but our understanding of this dependence remains tenuous.

Information theory has proven fruitful for understanding the competition between compression and relevance preservation in supervised learning~\cite{shwartz2017opening}.
We explored a similar trade-off in unsupervised learning, between multivariate information maximization and disentanglement of the learned factors. Writing this objective in terms of mutual information led to two surprising connections. First, we came to an unsupervised information bottleneck formulation that trades off compression and reconstruction relevance. Second, we found that by making appropriate variational approximations, we could reproduce the venerable VAE objective. This new perspective on VAE enabled more flexible distributions for latent codes and motivated new generalizations of the objective to localize interpretable information in latent codes. Ultimately, this led us to a novel learning objective that generated latent factors capturing intuitive structures in image data. We hope this alternative formulation of unsupervised learning continues to provide useful insights into this challenging problem. 

%% file: auto-corex.bbl
\begin{thebibliography}{34}
\providecommand{\natexlab}[1]{#1}
\providecommand{\url}[1]{\texttt{#1}}
\expandafter\ifx\csname urlstyle\endcsname\relax
  \providecommand{\doi}[1]{doi: #1}\else
  \providecommand{\doi}{doi: \begingroup \urlstyle{rm}\Url}\fi

\bibitem[Achille \& Soatto(2017)Achille and Soatto]{achille2017emergence}
Achille, Alessandro and Soatto, Stefano.
\newblock On the emergence of invariance and disentangling in deep
  representations.
\newblock \emph{arXiv preprint arXiv:1706.01350}, 2017.

\bibitem[Achille \& Soatto(2018)Achille and Soatto]{achille2018information}
Achille, Alessandro and Soatto, Stefano.
\newblock Information dropout: Learning optimal representations through noisy
  computation.
\newblock \emph{IEEE Transactions on Pattern Analysis and Machine
  Intelligence}, 2018.

\bibitem[Alemi et~al.(2017)Alemi, Fischer, Dillon, and Murphy]{alemi2016deep}
Alemi, Alexander~A, Fischer, Ian, Dillon, Joshua~V, and Murphy, Kevin.
\newblock Deep variational information bottleneck.
\newblock \emph{International Conference on Learning Representations}, 2017.

\bibitem[Barber \& Agakov(2003)Barber and Agakov]{barber2003algorithm}
Barber, David and Agakov, Felix.
\newblock The im algorithm: a variational approach to information maximization.
\newblock In \emph{Proceedings of the 16th International Conference on Neural
  Information Processing Systems}, pp.\  201--208. MIT Press, 2003.

\bibitem[Barlow(1989)]{barlow}
Barlow, Horace.
\newblock Unsupervised learning.
\newblock \emph{Neural computation}, 1\penalty0 (3):\penalty0 295--311, 1989.

\bibitem[Bengio et~al.(2017)Bengio, Thomas, Pineau, Precup, and
  Bengio]{bengio2017independently}
Bengio, Emmanuel, Thomas, Valentin, Pineau, Joelle, Precup, Doina, and Bengio,
  Yoshua.
\newblock Independently controllable features.
\newblock \emph{arXiv preprint arXiv:1703.07718}, 2017.

\bibitem[Bengio et~al.(2013)Bengio, Courville, and
  Vincent]{bengio2013representation}
Bengio, Yoshua, Courville, Aaron, and Vincent, Pascal.
\newblock Representation learning: A review and new perspectives.
\newblock \emph{IEEE transactions on pattern analysis and machine
  intelligence}, 35\penalty0 (8):\penalty0 1798--1828, 2013.

\bibitem[Chen et~al.(2016)Chen, Duan, Houthooft, Schulman, Sutskever, and
  Abbeel]{chen2016infogan}
Chen, Xi, Duan, Yan, Houthooft, Rein, Schulman, John, Sutskever, Ilya, and
  Abbeel, Pieter.
\newblock Infogan: Interpretable representation learning by information
  maximizing generative adversarial nets.
\newblock In \emph{Advances in Neural Information Processing Systems}, pp.\
  2172--2180, 2016.

\bibitem[Comon(1994)]{comon}
Comon, Pierre.
\newblock Independent component analysis, a new concept?
\newblock \emph{Signal processing}, 36\penalty0 (3):\penalty0 287--314, 1994.

\bibitem[Cover \& Thomas(2006)Cover and Thomas]{cover}
Cover, Thomas~M and Thomas, Joy~A.
\newblock \emph{Elements of information theory}.
\newblock Wiley-Interscience, 2006.

\bibitem[Dilokthanakul et~al.(2016)Dilokthanakul, Mediano, Garnelo, Lee,
  Salimbeni, Arulkumaran, and Shanahan]{dilokthanakul2016deep}
Dilokthanakul, Nat, Mediano, Pedro~AM, Garnelo, Marta, Lee, Matthew~CH,
  Salimbeni, Hugh, Arulkumaran, Kai, and Shanahan, Murray.
\newblock Deep unsupervised clustering with gaussian mixture variational
  autoencoders.
\newblock \emph{arXiv preprint arXiv:1611.02648}, 2016.

\bibitem[Dinh et~al.(2014)Dinh, Krueger, and Bengio]{nice}
Dinh, Laurent, Krueger, David, and Bengio, Yoshua.
\newblock Nice: Non-linear independent components estimation.
\newblock \emph{arXiv preprint arXiv:1410.8516}, 2014.

\bibitem[Goodfellow et~al.(2014)Goodfellow, Pouget-Abadie, Mirza, Xu,
  Warde-Farley, Ozair, Courville, and Bengio]{goodfellow2014generative}
Goodfellow, Ian, Pouget-Abadie, Jean, Mirza, Mehdi, Xu, Bing, Warde-Farley,
  David, Ozair, Sherjil, Courville, Aaron, and Bengio, Yoshua.
\newblock Generative adversarial nets.
\newblock In \emph{Advances in neural information processing systems}, pp.\
  2672--2680, 2014.

\bibitem[Gulrajani et~al.(2017)Gulrajani, Kumar, Ahmed, Taiga, Visin, Vazquez,
  and Courville]{gulrajani2016pixelvae}
Gulrajani, Ishaan, Kumar, Kundan, Ahmed, Faruk, Taiga, Adrien~Ali, Visin,
  Francesco, Vazquez, David, and Courville, Aaron.
\newblock Pixelvae: A latent variable model for natural images.
\newblock \emph{International Conference on Learning Representations}, 2017.

\bibitem[Higgins et~al.(2017)Higgins, Matthey, Pal, Burgess, Glorot, Botvinick,
  Mohamed, and Lerchner]{higgins2016beta}
Higgins, Irina, Matthey, Loic, Pal, Arka, Burgess, Christopher, Glorot, Xavier,
  Botvinick, Matthew, Mohamed, Shakir, and Lerchner, Alexander.
\newblock beta-vae: Learning basic visual concepts with a constrained
  variational framework.
\newblock In \emph{International Conference on Learning Representations}, 2017.

\bibitem[Kim \& Mnih(2017)Kim and Mnih]{kimdisentangling}
Kim, Hyunjik and Mnih, Andriy.
\newblock Disentangling by factorising.
\newblock \emph{NIPS Workshop on Learning Disentangled Representations}, 2017.

\bibitem[Kingma \& Welling(2013)Kingma and Welling]{kingma2013auto}
Kingma, Diederik~P and Welling, Max.
\newblock Auto-encoding variational bayes.
\newblock \emph{arXiv preprint arXiv:1312.6114}, 2013.

\bibitem[Linsker(1988)]{linsker}
Linsker, Ralph.
\newblock Self-organization in a perceptual network.
\newblock \emph{Computer}, 21\penalty0 (3):\penalty0 105--117, 1988.

\bibitem[Rezende et~al.(2014)Rezende, Mohamed, and
  Wierstra]{rezende2014stochastic}
Rezende, Danilo~Jimenez, Mohamed, Shakir, and Wierstra, Daan.
\newblock Stochastic backpropagation and approximate inference in deep
  generative models.
\newblock In \emph{International Conference on Machine Learning}, pp.\
  1278--1286, 2014.

\bibitem[Saxe et~al.(2018)Saxe, Bansal, Dapello, Advani, Kolchinsky, Daniel,
  and Cox]{saxe2018info}
Saxe, Michael~A, Bansal, Yamini, Dapello, Joel, Advani, Madhu, Kolchinsky,
  Artemy, Daniel, Brendan~T, and Cox, David~D.
\newblock On the information bottleneck theory of deep learning.
\newblock \emph{International Conference on Learning Representations}, 2018.

\bibitem[Schmidhuber(1992)]{factorial}
Schmidhuber, J{\"u}rgen.
\newblock Learning factorial codes by predictability minimization.
\newblock \emph{Neural Computation}, 4\penalty0 (6):\penalty0 863--879, 1992.

\bibitem[Shwartz-Ziv \& Tishby(2017)Shwartz-Ziv and Tishby]{shwartz2017opening}
Shwartz-Ziv, Ravid and Tishby, Naftali.
\newblock Opening the black box of deep neural networks via information.
\newblock \emph{arXiv preprint arXiv:1703.00810}, 2017.

\bibitem[S{\o}nderby et~al.(2016)S{\o}nderby, Raiko, Maal{\o}e, S{\o}nderby,
  and Winther]{sonderby2016ladder}
S{\o}nderby, Casper~Kaae, Raiko, Tapani, Maal{\o}e, Lars, S{\o}nderby,
  S{\o}ren~Kaae, and Winther, Ole.
\newblock Ladder variational autoencoders.
\newblock In \emph{Advances in neural information processing systems}, pp.\
  3738--3746, 2016.

\bibitem[Studen{\`y} \& Vejnarova(1998)Studen{\`y} and
  Vejnarova]{multiinformation}
Studen{\`y}, M and Vejnarova, J.
\newblock The multiinformation function as a tool for measuring stochastic
  dependence.
\newblock In \emph{Learning in graphical models}, pp.\  261--297. Springer,
  1998.

\bibitem[Thomas et~al.(2017)Thomas, Pondard, Bengio, Sarfati, Beaudoin, Meurs,
  Pineau, Precup, and Bengio]{thomas2017independently}
Thomas, Valentin, Pondard, Jules, Bengio, Emmanuel, Sarfati, Marc, Beaudoin,
  Philippe, Meurs, Marie-Jean, Pineau, Joelle, Precup, Doina, and Bengio,
  Yoshua.
\newblock Independently controllable features.
\newblock \emph{arXiv preprint arXiv:1708.01289}, 2017.

\bibitem[Tishby et~al.(2000)Tishby, Pereira, and Bialek]{tishby2000information}
Tishby, Naftali, Pereira, Fernando~C, and Bialek, William.
\newblock The information bottleneck method.
\newblock \emph{arXiv preprint physics/0004057}, 2000.

\bibitem[Van~Oord et~al.(2016)Van~Oord, Kalchbrenner, and
  Kavukcuoglu]{van2016pixel}
Van~Oord, Aaron, Kalchbrenner, Nal, and Kavukcuoglu, Koray.
\newblock Pixel recurrent neural networks.
\newblock In \emph{International Conference on Machine Learning}, pp.\
  1747--1756, 2016.

\bibitem[{Ver Steeg}(2017)]{steeg2017unsupervised}
{Ver Steeg}, Greg.
\newblock Unsupervised learning via total correlation explanation.
\newblock \emph{IJCAI}, 2017.

\bibitem[{Ver Steeg} \& Galstyan(2014){Ver Steeg} and
  Galstyan]{ver2014discovering}
{Ver Steeg}, Greg and Galstyan, Aram.
\newblock Discovering structure in high-dimensional data through correlation
  explanation.
\newblock In \emph{Advances in Neural Information Processing Systems}, pp.\
  577--585, 2014.

\bibitem[{Ver Steeg} \& Galstyan(2015){Ver Steeg} and
  Galstyan]{ver2015maximally}
{Ver Steeg}, Greg and Galstyan, Aram.
\newblock Maximally informative hierarchical representations of
  high-dimensional data.
\newblock In \emph{Artificial Intelligence and Statistics}, pp.\  1004--1012,
  2015.

\bibitem[{Ver Steeg} \& Galstyan(2017){Ver Steeg} and Galstyan]{steeg2017low}
{Ver Steeg}, Greg and Galstyan, Aram.
\newblock Low complexity gaussian latent factor models and a blessing of
  dimensionality.
\newblock \emph{arXiv preprint arXiv:1706.03353}, 2017.

\bibitem[Watanabe(1960)]{watanabe}
Watanabe, Satosi.
\newblock Information theoretical analysis of multivariate correlation.
\newblock \emph{IBM Journal of research and development}, 4\penalty0
  (1):\penalty0 66--82, 1960.

\bibitem[Zhao et~al.(2017{\natexlab{a}})Zhao, Song, and Ermon]{zhao2017infovae}
Zhao, Shengjia, Song, Jiaming, and Ermon, Stefano.
\newblock Infovae: Information maximizing variational autoencoders.
\newblock \emph{arXiv preprint arXiv:1706.02262}, 2017{\natexlab{a}}.

\bibitem[Zhao et~al.(2017{\natexlab{b}})Zhao, Song, and
  Ermon]{zhao2017learning}
Zhao, Shengjia, Song, Jiaming, and Ermon, Stefano.
\newblock Learning hierarchical features from generative models.
\newblock \emph{arXiv preprint arXiv:1702.08396}, 2017{\natexlab{b}}.

\end{thebibliography}
